\documentclass[10pt,twocolumn,letterpaper]{article}

\usepackage{iccv}
\usepackage{times}
\usepackage{epsfig}
\usepackage{graphicx}
\usepackage{amsmath}
\usepackage{amssymb}

\usepackage{latexsym}
\usepackage{amsmath}
\usepackage{amssymb}
\usepackage{xcolor}
\usepackage{enumerate}
\usepackage{comment}
\usepackage{subfigure}
\usepackage{multirow}
\usepackage{bm}
\usepackage{url}

\usepackage[breaklinks=true,bookmarks=false]{hyperref}

\iccvfinalcopy % *** Uncomment this line for the final submission

\def\iccvPaperID{3658} % *** Enter the ICCV Paper ID here

\ificcvfinal\fi

\begin{document}

%%%%%%%%% TITLE
\title{Multi-Angle Point Cloud-VAE: Unsupervised Feature Learning for 3D Point Clouds from Multiple Angles by Joint Self-Reconstruction and Half-to-Half Prediction}

\author{Zhizhong Han$^{1,3}$, Xiyang Wang$^{1,2}$, Yu-Shen Liu$^{1,2}$\thanks{Corresponding Author. This work was supported by National Key R\&D Program of China (2018YFB0505400) and NSF (award 1813583).}, Matthias Zwicker$^3$\\
%\affiliations
$^1$School of Software, Tsinghua University, Beijing, China \\
$^2$Beijing National Research Center for Information Science and Technology (BNRist)\\
$^3$Department of Computer Science, University of Maryland, College Park, USA\\
%\emails
{\tt\small h312h@umd.edu, wangxiya16@mails.tsinghua.edu.cn, liuyushen@tsinghua.edu.cn, zwicker@cs.umd.edu}
%\thanks{This work was supported by National Key R\&D Program of China (2018YFB0505400) and NSF under award number 1813583.}
}

\maketitle
\ificcvfinal\thispagestyle{empty}\fi

%%%%%%%%% ABSTRACT
\begin{abstract}
Unsupervised feature learning for point clouds has been vital for large-scale point cloud understanding. Recent deep learning based methods depend on learning global geometry from self-reconstruction. However, these methods are still suffering from ineffective learning of local geometry, which significantly limits the discriminability of learned features. To resolve this issue, we propose MAP-VAE to enable the learning of global and local geometry by jointly leveraging global and local self-supervision. To enable effective local self-supervision, we introduce multi-angle analysis for point clouds. In a multi-angle scenario, we first split a point cloud into a front half and a back half from each angle, and then, train MAP-VAE to learn to predict a back half sequence from the corresponding front half sequence. MAP-VAE performs this half-to-half prediction using RNN to simultaneously learn each local geometry and the spatial relationship among them. In addition, MAP-VAE also learns global geometry via self-reconstruction, where we employ a variational constraint to facilitate novel shape generation. The outperforming results in four shape analysis tasks show that MAP-VAE can learn more discriminative global or local features than the state-of-the-art methods.

\end{abstract}

%%%%%%%%% BODY TEXT
\section{Introduction}
Point clouds have become a popular 3D representation in machine vision, autonomous driving, and augmented reality, because they are easy to acquire and manipulate. Therefore, point cloud analysis has emerged as a crucial problem in the area of 3D shape understanding. With the help of extensive supervised information, recent deep learning based feature learning techniques have achieved unprecedented results in classification, detection and segmentation~\cite{nipspoint17,ShenFYT18,LiCL18,DGCNN2018,LiBSWDC18,XuFXZQ18}. However, supervised learning requires intense manual labeling effort to obtain supervised information. Therefore, unsupervised feature learning is an attractive alternative and a promising research challenge.

Several studies have tried to adress this challenge~\cite{PanosCVPR2018ICML,ChunLiangCVPR2019,eccvDengBI18,valsesia2018learning,YongbinCVPR2019,YonghengCVPR2019,Matan2019}. To learn the structure of a point cloud without additional supervision, these generative models are trained by self-supervision, such as self-reconstruction~\cite{PanosCVPR2018ICML,eccvDengBI18,mrt18,YonghengCVPR2019,YongbinCVPR2019,Matan2019} or distribution approximation~\cite{PanosCVPR2018ICML,ChunLiangCVPR2019,valsesia2018learning}, which is implemented by auto-encoder or generative adversarial networks~\cite{IanNIPS2014} respectively. To capture finer global structure, some methods~\cite{valsesia2018learning,YongbinCVPR2019,YonghengCVPR2019,Matan2019} first learn local structure information in point cloud patches based on which the global point cloud is then reconstructed. Because of lacking effective and semantic local structure supervision, however, error may accumulate in the local structure learning process, which limits the network's ability in 3D point cloud understanding.

To resolve this issue, we propose a novel deep learning model for unsupervised point cloud feature learning by simultaneously employing effective local and global self-supervision. We introduce multi-angle analysis for point clouds to mine effective local self-supervision, and combine it with global self-supervision under a variational constraint. Hence we call our model Multi-Angle Point Cloud Variational Auto-Encoder (MAP-VAE). Specifically, to employ local self-supervision, MAP-VAE first splits a point cloud into a front half and a back half under each of several incrementally varying angles. Then, MAP-VAE performs half-to-half prediction to infer a sequence of several back halves from the corresponding sequence of the complementary front halves. Half-to-half prediction aims to capture the geometric and structural information of local regions on the point cloud through varying angles. Moreover, by leveraging global self-supervision, MAP-VAE conducts self-reconstruction in company with each half-to-half prediction to capture the geometric and structural information of the whole point cloud. Self-reconstruction is started from a variational feature space, which enables MAP-VAE to generate new shapes by capturing the distribution information over training point clouds in the feature space. In summary, our contributions are as follows:

\begin{enumerate}[i)]
\item We propose MAP-VAE to perform unsupervised feature learning for point clouds. It can jointly leverage effective local and global self-supervision to learn fine-grained geometry and structure of point clouds.
\item We introduce multi-angle analysis for point cloud understanding, which provides semantic local self-supervision to learn local geometry and structure.
\item We provide a novel way to consistently split point clouds into semantic regions according to view angles, which enables the exploration of the fine-grained discriminative information of point cloud regions.
\end{enumerate}

\section{Related work}
Deep learning models have led to significant progress in feature learning for 3D shapes~\cite{Zhizhong2016b,Zhizhong2016,Han2017,HanTIP18,Zhizhong2018seq,Zhizhong2019seq,parts4features19,3DViewGraph19,3D2SeqViews19,HanCyber17a}. Here, we focus on reviewing studies on point clouds. For supervised methods, supervised information, such as shape class labels or segmentation labels, are required to train deep learning models in the feature learning process. In contrast, unsupervised methods are designed to mine self-supervision information from point clouds for training, which eliminates the need for supervised information that can be tedious to obtain. We briefly review the state-of-the-art methods in these two categories as follows.

\noindent \textbf{Supervised feature learning.} As a pioneering work, PointNet~\cite{cvprpoint2017} was proposed to directly learn features from point clouds by deep learning models. However, PointNet is limited in capturing contextual information among points. To resolve this issue, various techniques were proposed to establish graph in a local region to capture the relationship among points in the region~\cite{ShenFYT18,LiCL18,DGCNN2018,LiBSWDC18,XuFXZQ18}. Furthermore, multi-scale analysis~\cite{nipspoint17} was introduced to extract more semantic features from the local region by separating points into scales or bins, and then, aggregating these features by concatenation~\cite{XieLCT18} or RNN~\cite{p2seq18}. These methods require supervised information in the feature learning process, which is different from unsupervised approach in MAP-VAE.

%To train these methods, supervised information is required in the feature learning process, which is different from the unsupervised approach in our proposed MAP-VAE.

\noindent \textbf{Unsupervised feature learning.} An intuitive approach to mine self-supervised information is to perform self-reconstruction which first encodes a point cloud into a feature and then decodes the feature back to a point cloud. Such global self-supervision is usually implemented by an autoencoder network~\cite{PanosCVPR2018ICML,eccvDengBI18,YonghengCVPR2019,YongbinCVPR2019,Matan2019}. With the help of adversarial training, a different kind of global self-supervision is employed to train the network to generate plausible point clouds by learning a mapping from a known distribution to the unknown distribution that the point clouds are sampled from~\cite{PanosCVPR2018ICML,ChunLiangCVPR2019,valsesia2018learning}. For finer global structure information, some methods take a step further to jointly employ local structure information captured in local regions~\cite{valsesia2018learning,YongbinCVPR2019,YonghengCVPR2019,Matan2019}. These methods first learn local structure information in point cloud patches by clustering~\cite{YonghengCVPR2019}, conditional point distribution~\cite{YongbinCVPR2019}, graph convolution~\cite{valsesia2018learning}, or fully connected layers~\cite{Matan2019}, based on which the whole point cloud is then reconstructed. However, because of lacking effective and semantic local structure supervision, this process is prone to error accumulation in the local structure learning process, which limits the network's ability in point cloud understanding. To resolve this issue, MAP-VAE introduces multi-angle analysis for point clouds, which provides effective and semantic local self-supervision. MAP-VAE can also simultaneously employ local and global self-supervision, which further differentiates it from other methods.

\section{Overview}

\begin{figure}[tb]
  \centering
  % the following command controls the width of the embedded PS file
  % (relative to the width of the current column)
  %\includegraphics[width=.95\linewidth, bb=39 696 126 756]{figures/definition3.eps}
   \includegraphics[width=\linewidth]{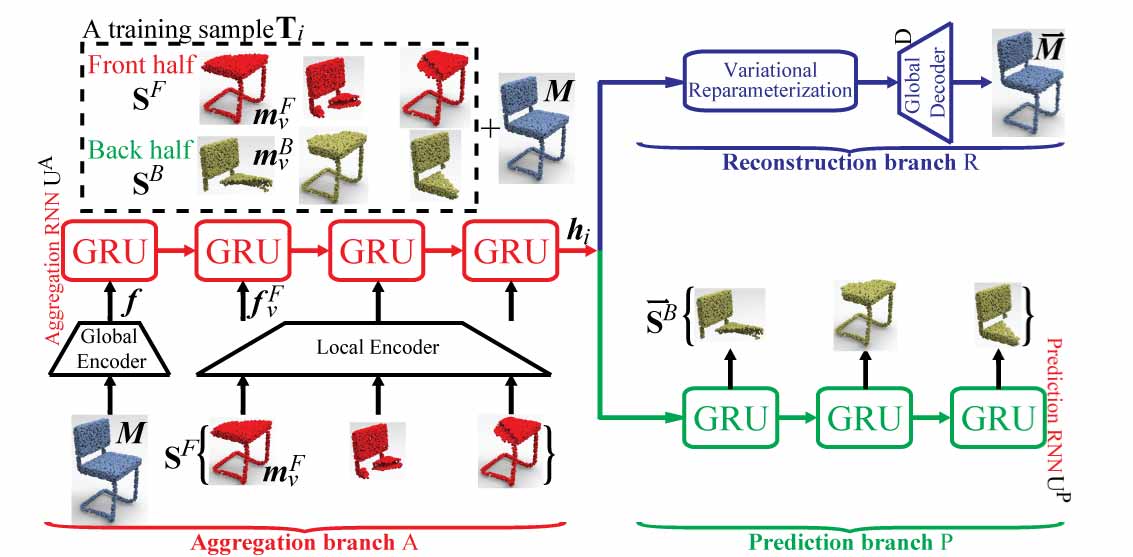}
  % replacing the above command with the one below will explicitly set
  % the bounding box of the PS figure to the rectangle (xl,yl),(xh,yh).
  % It will also prevent LaTeX from reading the PS file to determine
  % the bounding box (i.e., it will speed up the compilation process)
  % \includegraphics[width=.95\linewidth, bb=39 696 126 756]{sampleFig}
  %
  %
\caption{\label{fig:framework} The framework of MAP-VAE.
%To learn from a training sample $\bm{T}_i$, MAP-VAE conducts half-to-half prediction from the front half sequence $\mathbf{S}^F$ to the back half sequence $\mathbf{S}^B$ by aggregation branch $\rm A$ and prediction branch $\rm P$. Meanwhile, MAP-VAE conducts self-reconstruction of point cloud $\bm{M}$ by aggregation branch $\rm A$ and reconstruction branch $\rm R$.
}
\end{figure}

To jointly leverage local and global self-supervision to learn features from point clouds, MAP-VAE simultaneously conducts half-to-half prediction and self-reconstruction by three branches, i.e., which we call aggregation branch $\rm A$, reconstruction branch $\rm R$, and prediction branch $\rm P$, as illustrated in Fig.~\ref{fig:framework}. Specifically, branch $\rm A$ and branch $\rm P$ together perform the half-to-half prediction while branch $\rm A$ and branch $\rm R$ together perform the self-reconstruction.

A training sample $\mathbf{T}_i$ provided to MAP-VAE to learn is formed by a front half sequence $\mathbf{S}^F$, a back half sequence $\mathbf{S}^B$, and an original point cloud $\bm{M}$. The corresponding elements in sequences $\mathbf{S}^F$ and $\mathbf{S}^B$ are a front half $\bm{m}_v^F$ (in red) and its complementary back half $\bm{m}_v^B$ (in green) which are obtained by splitting the original point cloud $\bm{M}$ (in blue) from a specific angle $v$.

The aggregation branch $\rm A$ encodes the geometry of local point clouds and their spatial relationship by aggregating all front halves in sequence $\mathbf{S}^F$ in order. It first extracts the low-level feature $\bm{f}$ of the original point cloud $\bm{M}$ and the low-level feature $\bm{f}_v^F$ of each front half $\bm{m}_v^F$ by a global encoder and a local encoder, respectively. Then, it learns the angle-specific feature $\bm{h}_i$ of $\bm{M}$ by aggregating all low-level features $\bm{f}_v^F$ using an aggregation RNN $\rm U^A$.

The reconstruction branch $\rm R$ performs self-reconstruction by decoding the learned angle-specific feature $\bm{h}_i$ into a point cloud $\overline{\bm{M}}$. This reconstruction is conducted by a global decoder $\rm D$ which tries to generate $\overline{\bm{M}}$ as similar as possible to $\bm{M}$. In addition, $\rm R$ employs a variational constraint to facilitate novel shape generation.

At the same time, the prediction branch $\rm P$ performs half-to-half prediction by decoding the learned angle-specific feature $\bm{h}_i$ into a back half sequence $\overline{\mathbf{S}^{B}}$ which is paired with the corresponding front half sequence $\mathbf{S}^F$. This prediction is conducted by a prediction RNN $\rm U^P$ which tries to predict the sequence $\overline{\mathbf{S}^{B}}$ as similar as possible to $\mathbf{S}^B$.

\section{Multi-angle splitting}

\begin{figure}[tb]
  \centering
  % the following command controls the width of the embedded PS file
  % (relative to the width of the current column)
  %\includegraphics[width=.95\linewidth, bb=39 696 126 756]{figures/definition3.eps}
   \includegraphics[width=\linewidth]{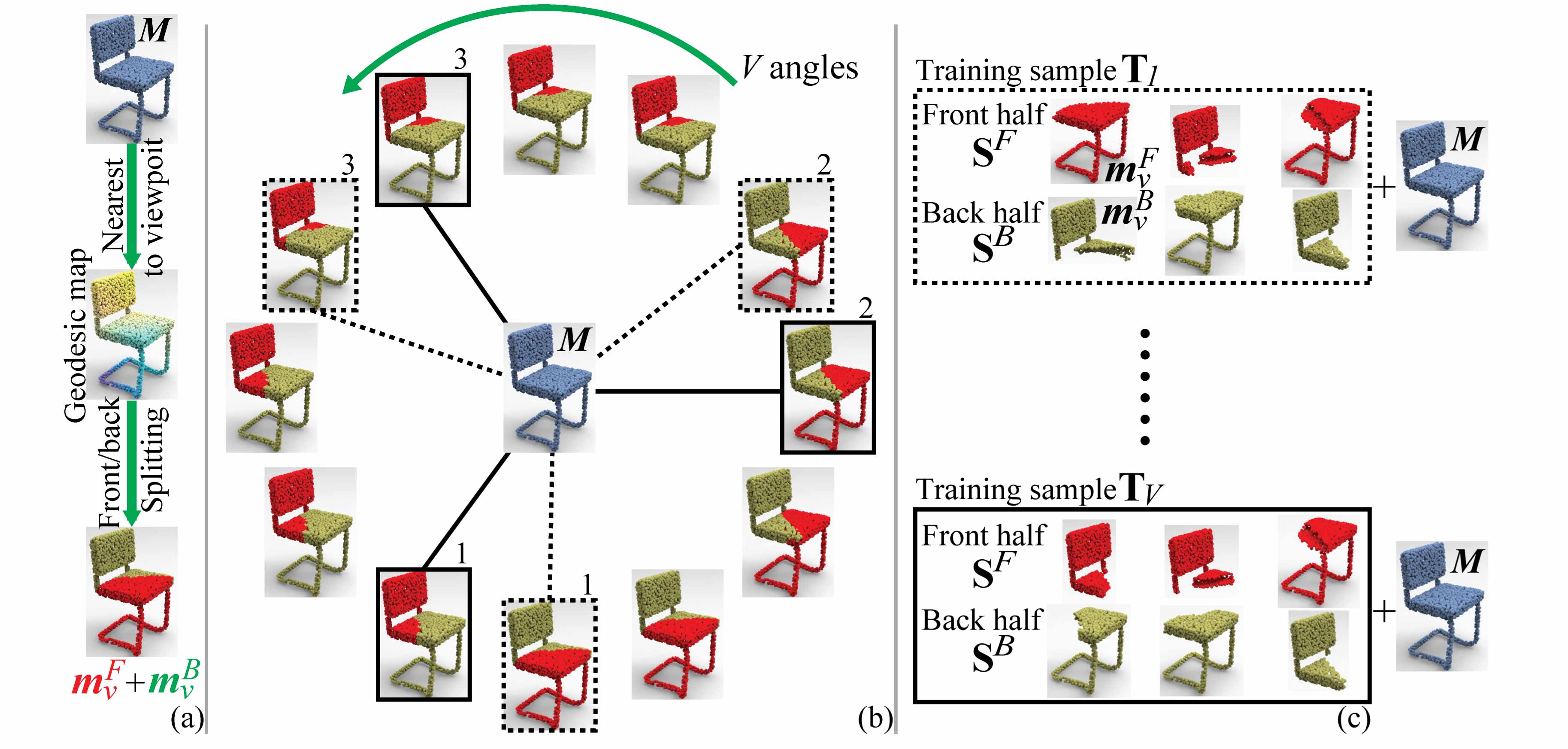}
  % replacing the above command with the one below will explicitly set
  % the bounding box of the PS figure to the rectangle (xl,yl),(xh,yh).
  % It will also prevent LaTeX from reading the PS file to determine
  % the bounding box (i.e., it will speed up the compilation process)
  % \includegraphics[width=.95\linewidth, bb=39 696 126 756]{sampleFig}
  %
  %
\caption{\label{fig:angles} %(a) The demonstration of geodesic splitting $\bm{M}$ from the $v$-th angle. (b) Establishing half-to-half sequence pairs (Each covers $W=3$ angles) based on geodesic splitting from all $V$ angles. (c) The training sample $\mathbf{T}_i$ from each one of $V$ angles.
%(a) A point cloud $\bm{M}$ is split into a front half $\bm{m}_v^F$ (in red) and a back half $\bm{m}_v^B$ (in green) from a given angle in terms of geodesic distance on $\bm{M}$. (b) $\bm{M}$ is further split into different front and back halves from $V$ angles located around $\bm{M}$ in clockwise order, where subset with $W=3$ out of $V$ angles (indicated by dotted or solid line) are selected to establish a half-to-half sequence pair. (c) Starting with each one of $V$ angles, we obtain a training sample $\mathbf{T}_i$.
(a) Geodesic splitting $\bm{M}$ into a front half $\bm{m}_v^F$ (in red) and a back half $\bm{m}_v^B$ (in green) from the $v$-th angle. (b) $\bm{M}$ is further split from all $V$ angles located around $\bm{M}$ in clockwise order, where subset with $W=3$ out of $V$ angles (indicated by dotted or solid line) are selected to establish a half-to-half sequence pair. (c) The training sample $\mathbf{T}_i$ from each one of $V$ angles.
}
\end{figure}

\noindent \textbf{Multi-angle splitting.} A key idea in MAP-VAE is a novel multi-angle analysis for point coulds to mine effective local self-supervision. Intuitively, observing a point cloud from different angles, explicitly presents the correspondences and relationships among different shape regions, given as the correspondence between the front and back halves of the shape in each view. Our multi-angle analysis provides multiple regions (front halves) of the point cloud as input, from which the corresponding missing regions (back halves) need to be inferred. This encourages MAP-VAE to learn a detailed shape representation that facilitates high quality classification, segmentation, shape synthesis, and point cloud completion.

We achieve this by splitting a point cloud into a front and a back half from different angles, where the front half is the half nearer to the viewpoint than the back half. This enables MAP-VAE to observe different semantic parts of a point cloud, and it also preserves the spatial relationship among the parts by incrementally varying angles.

For a point cloud $\bm{M}$, we split $\bm{M}$ from $V$ different angles into front halves (in red) and their complementary back halves (in green), as shown in Fig.~\ref{fig:angles} (b), where the viewpoints are located around $\bm{M}$ on a circle. From the $v$-th viewpoint, $\bm{M}$ is split into a front half $\bm{m}_v^F$ and a back half $\bm{m}_v^B$ in Fig.~\ref{fig:angles} (a), where $\bm{m}_v^F$ is formed by the $N$ nearest points (in red) of $\bm{M}$ to the viewpoint while $\bm{m}_v^B$ is formed by the $N$ farthest points (in green) to the viewpoint.

\begin{figure}[tb]
  \centering
  % the following command controls the width of the embedded PS file
  % (relative to the width of the current column)
  %\includegraphics[width=.95\linewidth, bb=39 696 126 756]{figures/definition3.eps}
   \includegraphics[width=\linewidth]{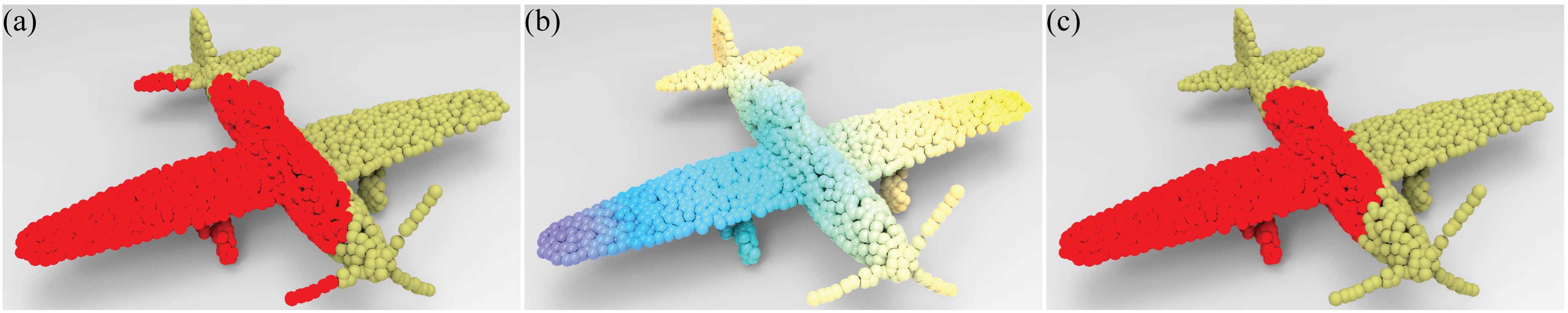}
  % replacing the above command with the one below will explicitly set
  % the bounding box of the PS figure to the rectangle (xl,yl),(xh,yh).
  % It will also prevent LaTeX from reading the PS file to determine
  % the bounding box (i.e., it will speed up the compilation process)
  % \includegraphics[width=.95\linewidth, bb=39 696 126 756]{sampleFig}
  %
  %
\caption{\label{fig:ordering} The comparison of front halves (red parts in (a) and (c)) split by Euclidean distance and geodesic distance (map in (b)).
%The front half (red in (a)) split by Euclidean distance is not continuous, while geodesic distance (map in (b)) splits a contiguous one (red in (a)).
%We compare different splitting methods. (a) With Euclidean distance, the obtained front half (in red) is not continuous. (c) Our method leverages the geodesic distance (b) to obtain a continuous front half (in red).
}
\end{figure}

\noindent\textbf{Geodesic splitting. }A naive way of finding the $N$ nearest points to define a front half $\bm{m}_v^F$ is to sort all points on $\bm{M}$ by the Euclidean distance between each point and the viewpoint. However, on some point clouds, this method may not produce semantic front halves, since the regions in a front half are not continuous, as demonstrated in Fig.~\ref{fig:ordering} (a). It is important to encode semantic front halves, since this would help MAP-VAE to seamlessly understand the entire surface from a viewpoint under a multi-angle scenario.

To resolve this issue, we leverage the geodesic distance on the point cloud~\cite{Crane:2017:HMD} to sort the points. Specifically, we first find a nearest point $\bm{u}$ to the $v$-th viewpoint on $\bm{M}$ by Euclidean distance. Then, we sort the rest of points on $\bm{M}$ in terms of their geodesic distances to the found nearest point $\bm{u}$, as shown by the geodesic map in Fig.~\ref{fig:ordering} (b). Finally, $\bm{u}$ and its nearest $N-1$ points form the front half $\bm{m}_v^F$, while the farthest $N$ points form the back half $\bm{m}_v^B$, as illustrated by the red part and green part in Fig.~\ref{fig:ordering} (c), respectively.

\noindent \textbf{Half-to-half sequence pairs.} To leverage the correspondence between front half and back half and their spatial relationship under different angles, we establish a half-to-half sequence pair starting from each one of the $V$ angles.

Along the circle direction of varying angles, as illustrated by the clockwise green arrow in Fig.~\ref{fig:angles} (b), we select front halves $\bm{m}_v^F$ and their complementary back halves $\bm{m}_v^B$ from $W$ out of the $V$ angles. The selected $\bm{m}_v^F$ form a front half sequence $\mathbf{S}^F$ while the complementary $\bm{m}_v^B$ form a back half sequence $\mathbf{S}^B$, where $\mathbf{S}^F=[\bm{m}_v^F|v\in[1,V], \lvert v \rvert=W]$, $\mathbf{S}^B=[\bm{m}_v^B|v\in[1,V], \lvert v \rvert=W]$ and each element in $\mathbf{S}^F$ corresponds to its complementary element in $\mathbf{S}^B$. Thus, a half-to-half sequence pair $(\mathbf{S}^F,\mathbf{S}^B)$ consists of $\mathbf{S}^F$ and $\mathbf{S}^B$.

To comprehensively observe the point cloud, we select $W$ angles which uniformly cover the whole shape in each half-to-half sequence pair $(\mathbf{S}^F,\mathbf{S}^B)$. As demonstrated by the dotted lines in Fig.~\ref{fig:angles} (b), we select $W=3$ angles in order to form the first $(\mathbf{S}^F,\mathbf{S}^B)$, and then, along the green arrow, we form the last $(\mathbf{S}^F,\mathbf{S}^B)$ by angles indicated by the solid lines. Each $(\mathbf{S}^F,\mathbf{S}^B)$ forms a training sample $\mathbf{T}_i$ in company with the point cloud $\bm{M}$. Finally, we obtain all $V$ training samples $\{\mathbf{T}_i|i\in[1,V]\}$ from $\bm{M}$ in Fig.~\ref{fig:angles} (c).

\section{MAP-VAE}
\noindent \textbf{Aggregation branch $\rm A$.} For a training sample $\mathbf{T}_i$ containing a half-to-half sequence pair $(\mathbf{S}^F,\mathbf{S}^B)$ and the point cloud $\bm{M}$, aggregation branch $\rm A$ encodes the global geometry of $\bm{M}$, local geometry of each one of $W$ $\mathbf{m}_v^F$ in $\mathbf{S}^F$, and the spatial relationship among $\mathbf{m}_v^F$. Aggregation branch $\rm A$ first extracts the geometry of each involved point cloud into a low-level feature by encoder, and then, aggregates all the low-level features with their spatiality by aggregation RNN $\rm U^A$. Specifically, we extract the low-level feature $\bm{f}$ of $2N$ points on $\bm{M}$ by a global encoder, and the low-level feature $\bm{f}_v^F$ of $N$ points on $\mathbf{m}_v^F$ by a local encoder. Both the global and local encoders employ the same architecture as the encoder in PointNet++~\cite{nipspoint17}, the only difference is the input number of points. Subsequently, aggregation RNN $\rm U^A$ aggregates $\bm{f}$ and all $\bm{f}_v^F$ in $W+1$ steps, where we employ GRU cell with $512$ hidden state. Finally, we use the hidden state as the angle-specific feature $\bm{h}_i$ of $\bm{M}$ since the first front half in $\mathbf{S}^F$ is observed starting from the $i$-th angle.

\noindent \textbf{Reconstruction branch $\rm R$.} By decoding the learned angle-specific feature $\bm{h}_i$, reconstruction branch $\rm R$ tries to generate a point cloud $\overline{\bm{M}}$ as similar as possible to the original point cloud $\bm{M}$ by a global decoder $\rm D$. $\rm D$ is formed by 3 fully connected layers (1024-2048-6114) and 2 convolutional layers (with 256 and 3 $1\times 1$ kernels each), where batch normalization is used between every two layers. Here, we prefer Earth Mover¡¯s distance (EMD)~\cite{Rubner2000} to Chamfer Distance (CD)~\cite{FanSG17} to evaluate the distance between the reconstructed $\overline{\bm{M}}$ and the original $\bm{M}$, since EMD is more faithful than CD to the visual quality of point clouds\cite{PanosCVPR2018ICML}. The EMD distance between $\overline{\bm{M}}$ and $\bm{M}$ is regarded as the cost of reconstruction to optimize, as defined below, where $\phi$ is a bijection from a point $x$ on $\bm{M}$ to its corresponding point $\phi(x)$ on $\overline{\bm{M}}$,

%\begin{equation}
%\label{eq:cd}
%C_{\rm D}=\sum_{x\in\bm{M}} \min_{y\in\overline{\bm{M}}}\Vert x-y\Vert_2^2+\sum_{y\in\overline{\bm{M}}} \min_{x\in\bm{M}}\Vert x-y\Vert_2^2.
%\end{equation}

\begin{equation}
\label{eq:emd}
C_{\rm D}=\min_{\phi:\bm{M}\to\overline{\bm{M}}}\sum_{x\in\bm{M}}\Vert x-\phi(x)\Vert_2.
\end{equation}

In addition, we employ a variational constraint~\cite{KingmaW13} in reconstruction branch $\rm R$ to facilitate novel shape generation. This is implemented by a variational reparameterization process, as shown in Fig.~\ref{fig:framework}. The variational reparameterization transforms the angle-specific feature $\bm{h}_i$ into another latent vector $\bm{z}$ that roughly follows a unit multi-dimensional Gaussian distribution. After training, branch $\rm R$ can generate a novel shape by sampling a latent vector from the unit Gaussian to the global decoder $\rm D$.

Specifically, the variational reparameterization first employs fully connected layers to respectively estimate the mean $\bm{\mu}$ and variance $\bm{\sigma}$ for the distribution of $\bm{h}_i$. Then, a noise vector $\bm{\varepsilon}$ is sampled from a unit multi-dimensional Gaussian distribution $\mathcal{N}(\bm{0},\bm{1})$, as $\bm{\varepsilon}\sim\mathcal{N}(\bm{0},\bm{1})$ and $\bm{\varepsilon}\in {\rm R}^{1\times Z}$. Finally, we scale the noise $\bm{\varepsilon}$ by $\bm{\sigma}$ and further shift it by $\bm{\mu}$, such that the latent vector $\bm{z}=\bm{\mu}+\bm{\varepsilon}\odot\bm{\sigma}$.
The variational reparameterization enables reconstruction branch $\rm R$ to push the distribution $q(\bm{z}|\bm{h}_i)$ to follow the unit multi-dimensional Gaussian distribution by minimizing the KL divergence between the distribution $q(\bm{z}|\bm{h}_i)$ and $\mathcal{N}(\bm{0},\bm{1})$.

Thus, the cost of reconstruction branch $\rm R$ is defined based on Eq.~(\ref{eq:emd}) below, where $\alpha$ is a balance parameter.

\begin{equation}
\label{eq:kl}
C_{\rm R}=C_{\rm D}+\alpha \times {\rm KL} (q(\bm{z}|\bm{h}_i)||\mathcal{N}(\bm{0},\bm{1})).
\end{equation}

\noindent \textbf{Prediction branch {\rm P}.} Similar to reconstruction branch $\rm R$, prediction branch {\rm P} decodes the learned angle-specific feature $\bm{h}_i$ to predict the back half sequence $\mathbf{S}^B$ corresponding to $\mathbf{S}^F$. Branch $\rm P$ tries to predict a back half sequence $\overline{\mathbf{S}^B}$ as similar as possible to $\mathbf{S}^B$ by a prediction RNN $\rm U^P$. At each of $W$ steps, $\rm U^P$ predicts one back half $\overline{\mathbf{m}_v^B}$ in the same order of elements in $\mathbf{S}^B$. This enables $\rm U^P$ to learn the half-to-half correspondence and the spatial relationship among the halves of $\bm{M}$. To further push MAP-VAE to comprehensively understand the point cloud, $\rm U^P$ predicts the low-level feature $\overline{\bm{f}_v^B}$ of each one of $W$ back half $\overline{\mathbf{m}_v^B}$ rather than the spatial point coordinates of $\overline{\mathbf{m}_v^B}$, which is complementary to reconstruction branch $\rm R$. The ground truth low-level feature $\bm{f}_v^B$ of $\mathbf{m}_v^B$ is also extracted by the local encoder in branch $\rm A$. Thus, the cost of branch $\rm P$ is defined as follows,

\begin{equation}
\label{eq:prediction}
C_{\rm P}=\frac{1}{W}\times\sum_{v\in [1,V], \lvert v\rvert=W} \Vert \overline{\bm{f}_v^B}-\bm{f}_v^B \Vert_2^2.
%C_{\rm P}=\frac{1}{W}\times\sum_{\overline{\bm{f}_v^B}\in \overline{\mathbf{S}^B},\bm{f}_v^B\in \mathbf{S}^B} \Vert \overline{\bm{f}_v^B}-\bm{f}_v^B \Vert_2^2
\end{equation}

\noindent \textbf{Objective function.} For a sample $\mathbf{T}_i$, MAP-VAE is trained by minimizing all the aforementioned costs of each branch, as defined below, where $\beta$ is a balance parameter.

\begin{equation}
\label{eq:prediction}
\min \ C_{\rm R}+\beta\times C_{\rm P}.
\end{equation}

After training, MAP-VAE represents the point cloud $\bm{M}$ as a global feature $\bm{H}$ by aggregating the angle-specific feature $\bm{h}_i$ learned from each sample $\mathbf{T}_i$ of $\bm{M}$ using max pooling, such that $\bm{H}={\rm Pool}_{i\in[1,V]}\{\bm{h}_i\}$.

\section{Experimental results and analysis}
In this section, we first explore how the parameters involved in MAP-VAE affect the discriminability of learned global features in shape classification. Then, MAP-VAE is evaluated in shape classification, segmentation, novel shape generation, and point cloud completion by comparing with state-of-the-art methods.

\noindent \textbf{Training.} We pre-train the global and local encoders in MAP-VAE respectively under the dataset involved in experiments in a self-reconstruction task, where the decoder in PointNet++~\cite{nipspoint17} for segmentation is modified to work with our encoders to produce three dimensional point coordinates in the last layer. After each PointNet++-based autoencoder is trained, the pre-trained global and local encoders are fixed for more efficient training of MAP-VAE.

\begin{figure}[tb]
  \centering
  % the following command controls the width of the embedded PS file
  % (relative to the width of the current column)
  %\includegraphics[width=.95\linewidth, bb=39 696 126 756]{figures/definition3.eps}
   \includegraphics[width=\linewidth]{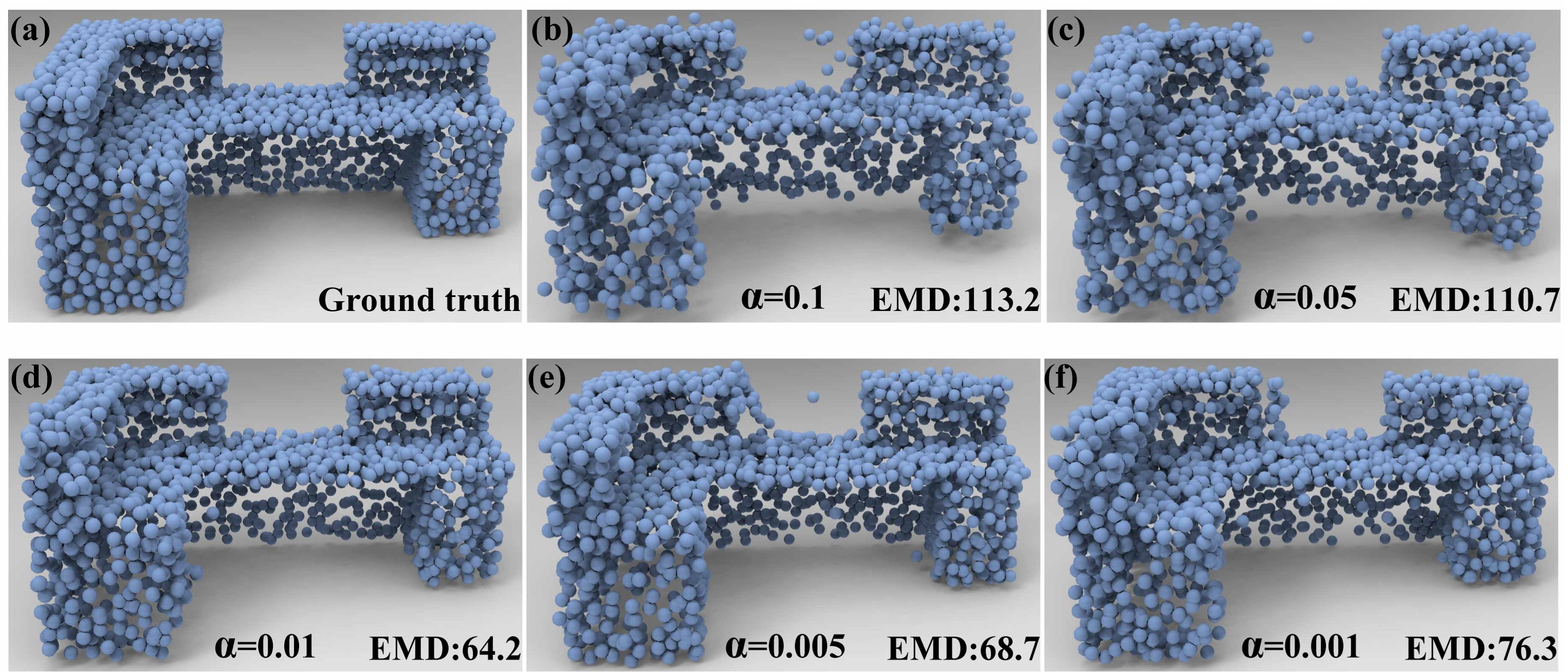}
  % replacing the above command with the one below will explicitly set
  % the bounding box of the PS figure to the rectangle (xl,yl),(xh,yh).
  % It will also prevent LaTeX from reading the PS file to determine
  % the bounding box (i.e., it will speed up the compilation process)
  % \includegraphics[width=.95\linewidth, bb=39 696 126 756]{sampleFig}
  %
  %
\caption{\label{fig:KLweight} The point clouds are reconstructed in (b)-(f) under different $\alpha$ compared in Table.~\ref{table:balanceAlpha}.
%The effect of reconstruction is affected by $\alpha$ in a similar way shown in Table.~\ref{table:balanceAlpha} in terms of the EMD distance to the ground truth in (a).
}
\end{figure}

In all experiments, we choose a more challenging way to train MAP-VAE by all point clouds in multiple shape classes of a benchmark rather than a single shape class, where each point cloud has $2048$ points and each half has $N=1024$ points. In shape classification experiments, we train a linear SVM to evaluate the raw discriminability of the learned global feature $\bm{H}$.

\begin{figure}[tb]
  \centering
  % the following command controls the width of the embedded PS file
  % (relative to the width of the current column)
  %\includegraphics[width=.95\linewidth, bb=39 696 126 756]{figures/definition3.eps}
   \includegraphics[width=\linewidth]{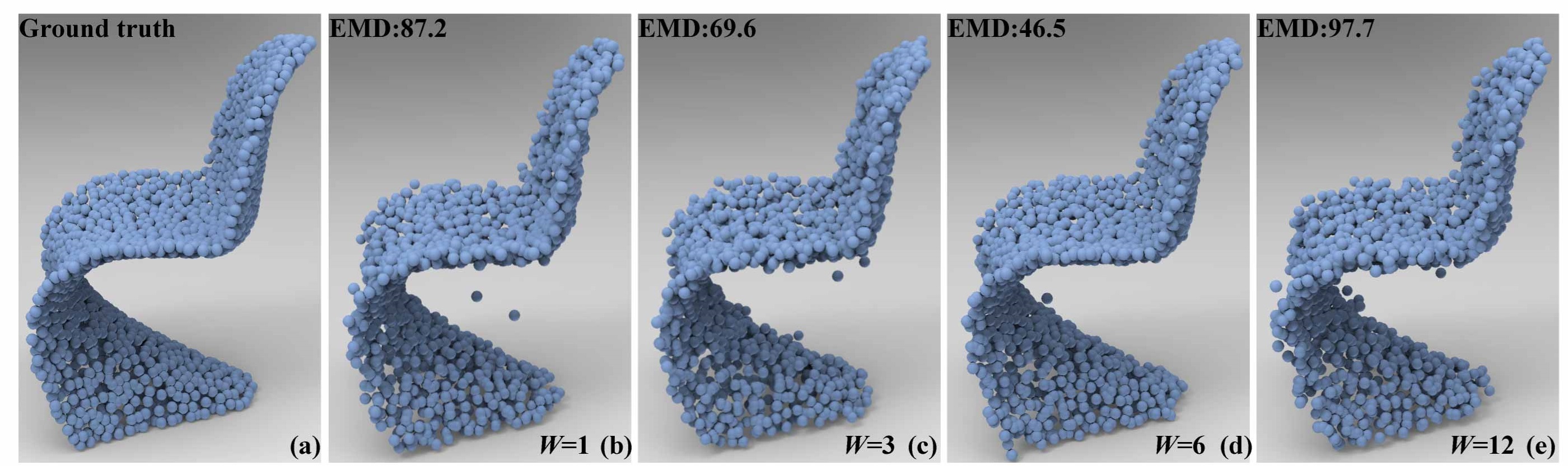}
  % replacing the above command with the one below will explicitly set
  % the bounding box of the PS figure to the rectangle (xl,yl),(xh,yh).
  % It will also prevent LaTeX from reading the PS file to determine
  % the bounding box (i.e., it will speed up the compilation process)
  % \includegraphics[width=.95\linewidth, bb=39 696 126 756]{sampleFig}
  %
  %
\caption{\label{fig:WCompare} The original point clouds in (a) are reconstructed in (b)-(e) under different $W$ compared in Table.~\ref{table:balanceW}.
%The effect of reconstruction is affected by $W$ in a similar way shown in Table.~\ref{table:balanceW} in terms of the EMD distance to the ground truth in (a).
}
\end{figure}

Initially, we employ $V=12$ angles to analyze a point cloud and form a training sample by $W=6$ angles uniformly covering the point cloud. We set balance parameter $\alpha=0.01$ and $\beta=1000$ to make each cost in the same order of magnitude. We use a $Z=128$ dimensional unit Gaussian for the variational constraint.

\noindent \textbf{Parameter setting.} All experiments on parameter effect exploration are conducted under ModelNet10~\cite{Wu2015}.

We first evaluate how $\beta$ affects MAP-VAE by comparing the results of different $\beta$ candidates including $\{10,100,1000,10000\}$. As shown in Table~\ref{table:balanceBeta}, the results get better with increasing $\beta$ until $\beta=1000$ and degenerate when $\beta$ is too big. This observation demonstrates a proper range of $\beta$. We use $\beta=1000$ in the following experiments.

\begin{table}
  \caption{The effect of $\beta$, $\alpha=0.01$, $W=6$, $Z=128$.}%$\varepsilon=0.0004$,
  \label{table:balanceBeta}
  \centering
  \begin{tabular}{c|cccc}%llllllllll
    \hline
     $\beta$ &10&100&1000&10000\\
    \hline		
     ACC$\%$ & 93.72 & 93.94 & \textbf{94.82} & 93.72\\
    \hline
  \end{tabular}
\end{table}

Then, we evaluate how $\alpha$ affects MAP-VAE by comparing the results of different $\alpha$ candidates including $\{0.1,0.05,0.01,0.005,0.001\}$. As shown in Table~\ref{table:balanceAlpha}, the results get better with decreasing $\alpha$ until $\alpha=0.01$ and degenerate when $\alpha$ is too small. This observation demonstrates how enforcing a unit Gaussian distribution on the latent vector $\bm{z}$ too loosely or strictly affects the discriminability of learned global features. We also visualize the point clouds reconstructed by branch $\rm R$ under different $\alpha$, as demonstrated in Fig.~\ref{fig:KLweight}. We find $\alpha$ affects the reconstructed point clouds in a similar way to how it affects the discriminability of learned global features. In the following experiments, we set $\alpha$ to 0.01.

\begin{table}
  \caption{The effect of $\alpha$, $\beta=1000$, $W=6$, $Z=128$.}%$\varepsilon=0.0004$,
  \label{table:balanceAlpha}
  \centering
  \begin{tabular}{c|ccccc}%llllllllll
    \hline
     $\alpha$ &0.1&0.05&0.01&0.005&0.001\\
    \hline		
     ACC$\%$ & 92.62 & 92.84 & \textbf{94.82} & 93.39&93.17\\
    \hline
  \end{tabular}
\end{table}

Subsequently, we explore how the number of angles $W$ of in a training sample affects the performance of MAP-VAE, as shown in Table.~\ref{table:balanceW}, where several candidate $W$ including $\{1,3,6,12\}$ are employed. We find $W=6$ achieves the best result, where fewer angles provide less local information while more angles increase redundancy. We also observe a similar phenomenon in the reconstructed point clouds shown in Fig.\ref{fig:WCompare}. In addition, we also explore other ways of distributing the $W=6$ angles, such as continuously (``S-6'') or randomly (``R-6''), respectively. We find our employed uniform placement is the best, since each training sample could cover the whole point cloud. In the following experiments, we use $W=6$.

\begin{table}
  \caption{The effect of $W$, $\alpha=0.01$, $\beta=1000$, $Z=128$.}%$\varepsilon=0.0004$,
  \label{table:balanceW}
  \centering
  \begin{tabular}{c|cccccc}%llllllllll
    \hline
     $W$ &1&3&6&12&S-6&R-6\\
    \hline		
     ACC$\%$ & 92.95 & 93.39 & \textbf{94.82} & 93.17&93.39&92.95\\
    \hline
  \end{tabular}
\end{table}

Finally, we explore the effect of the $Z$-dimensional unit Gaussian distribution. In Table.~\ref{table:balanceZ}, we compare the results obtained with different $Z$, including $\{32,64,128,256\}$. The results get better with increasing $Z$ until $Z=128$ while degenerating when $Z$ is too big. We believe $Z$ depends on the number of training samples, and both $64$ and $128$ are good for $Z$ under ModelNet10. $Z$ is set to 128 below.

\begin{table}
  \caption{The effect of $Z$, $\alpha=0.01$, $\beta=1000$, $W=6$.}%$\varepsilon=0.0004$,
  \label{table:balanceZ}
  \centering
  \begin{tabular}{c|cccc}%llllllllll
    \hline
     $Z$ &32&64&128&256\\
    \hline		
     ACC$\%$ & 93.28 & 94.16 & \textbf{94.82} & 93.94\\
    \hline
  \end{tabular}
\end{table}

\noindent \textbf{Ablation study.} We further explore how each module in MAP-VAE contributes to the performance. As shown in Table.~\ref{table:balanceStudy}, we remove a loss each time to highlight the corresponding module. The degenerated results indicate that all elements contribute to the discriminability of learned features, and self-reconstruction (``No $\rm R$'') contributes more than the half-to-half prediction (``No $\rm P$'').

\begin{figure*}[tb]
  \centering
  % the following command controls the width of the embedded PS file
  % (relative to the width of the current column)
  %\includegraphics[width=.95\linewidth, bb=39 696 126 756]{figures/definition3.eps}
   \includegraphics[width=\linewidth]{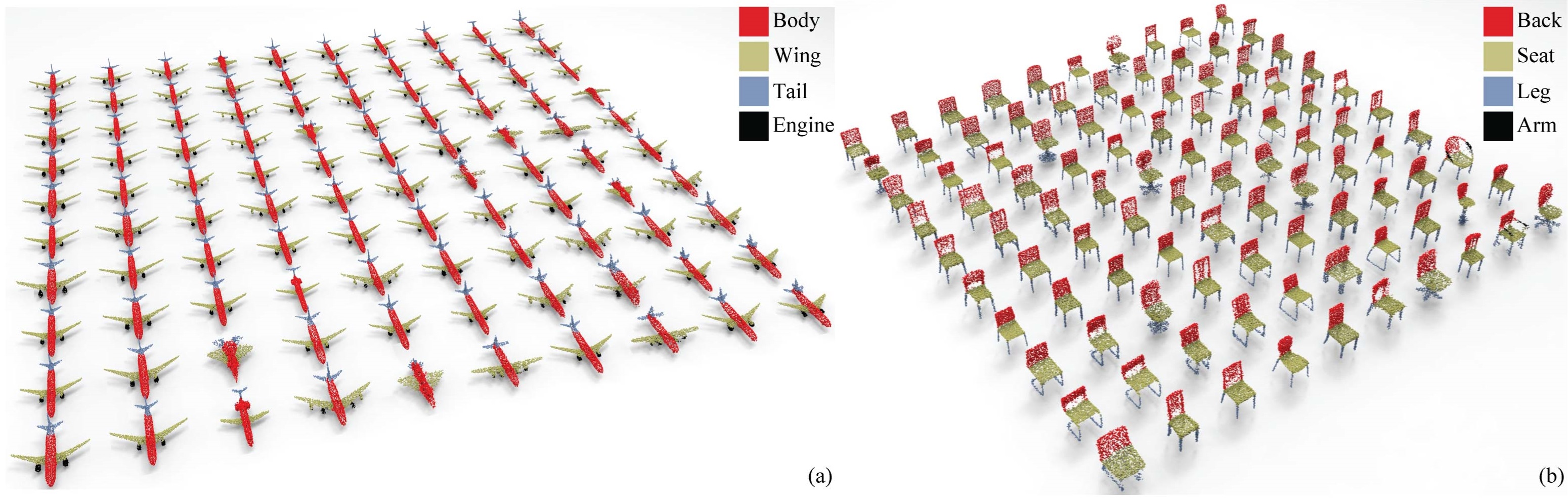}
  % replacing the above command with the one below will explicitly set
  % the bounding box of the PS figure to the rectangle (xl,yl),(xh,yh).
  % It will also prevent LaTeX from reading the PS file to determine
  % the bounding box (i.e., it will speed up the compilation process)
  % \includegraphics[width=.95\linewidth, bb=39 696 126 756]{sampleFig}
  %
  %
\caption{\label{fig:Seg} We show segmentation results from the airplane class in (a) and the chair class in (b).}
\end{figure*}

In addition, we highlight our half-to-half prediction by showing the results obtained only by the pre-trained global encoder in Fig.~\ref{fig:framework} and this global encoder with a variational constraint (using the same balance weights as MAP-VAE), as shown by ``AE'' and ``VAE''. These results show that half-to-half prediction can help MAP-VAE understand point clouds better by leveraging effective local self-supervision. Moreover, the lower result of ``Euclid'' also indicates geodesic distance is superior to Euclidean distance to obtain semantic front half in the splitting of a point cloud.

\begin{table}
  \caption{Ablation study,$\alpha=0.01$,$\beta=1000$,$W=6$,$Z=128$.}%$\varepsilon=0.0004$,
  \label{table:balanceStudy}
  \centering
   \resizebox{0.5\textwidth}{!}{
  \begin{tabular}{c|ccccccc}%llllllllll
    \hline
      &No {\rm R}&No {\rm P}&No {\rm KL}& All &AE&VAE&Eucli\\
    \hline		
     $\%$ & 91.63 & 92.40 & 93.17 & \textbf{94.82}&92.29&93.28&93.61\\
    \hline % no  R 91.629 % AE 92.293 % VAE 93.282 Eul 0.9361233
  \end{tabular}}
\end{table}

\noindent \textbf{Shape classification.} We evaluate MAP-VAE in shape classification by comparing it with state-of-the-art methods under ModelNet40~\cite{Wu2015} and ModelNet10~\cite{Wu2015}. All the compared methods perform unsupervised 3D feature learning while using various 3D representations, including voxels, views and point clouds. As shown in Table~\ref{table:comparison}, MAP-VAE obtains the best performance among these methods under ModelNet10. We employ the same parameters involved in Table~\ref{table:balanceStudy} to produce our result under ModelNet40, where MAP-VAE also outperforms all point cloud based methods. Although view-based VIPGAN is a little better than ours, it cannot generate 3D shapes. These results indicate that MAP-VAE learns more discriminative global features for point clouds with the ability of leveraging more effective local self-supervision. Note that the results of LGAN, FNet and NSampler are trained under a version of ShapeNet55 that contains more than 57,000 3D shapes. However, there are only 51,679 3D shapes from ShapeNet55 that are available for public download. Therefore, MAP-VAE cannot be trained under the same number of shapes. To perform fair comparison, we use the codes of LGAN and FNet to produce their results under the same shapes in ModelNet, as shown by ``LGAN(MN)'' and ``FNet(MN)''.
%, which employs the same training data as ours and other methods.

\begin{table}
  \caption{The classification accuracy ($\%$) comparison among unsupervised 3D feature learning methods under ModelNet40 and ModelNet10. $\alpha=0.01$,$\beta=1000$,$W=6$,$Z=128$.}
  \label{table:comparison}
  \centering
  \begin{tabular}{cccc}
    \hline
    Methods & Modality  & MN40$\%$ & MN10$\%$ \\
    \hline
    T-L Network\cite{Girdhar16} & Voxel & 74.40 & - \\
    Vconv-DAE\cite{Sharma16} & Voxel & 75.50 & 80.50\\
    3DGAN\cite{WuNIPS2016} & Voxel & 83.30 &91.00\\
    VSL\cite{liu2018learning} & Voxel & 84.50 & 91.00 \\
    VIPGAN\cite{Zhizhong2018VIP}& View &91.98&94.05\\
    LGAN\cite{PanosCVPR2018ICML} & Points & 85.70&95.30\\
    LGAN\cite{PanosCVPR2018ICML}(MN) & Points & 87.27&92.18\\
    NSampler\cite{Edoardo19} & Points & 88.70 & 95.30 \\
    FNet\cite{YaoqingCVPR2018} & Points& 88.40 &94.40\\
    FNet\cite{YaoqingCVPR2018}(MN) & Points& 84.36 &91.85\\
    MRTNet\cite{mrt18} & Points & 86.40 & - \\
    %NSampler\cite{Edoardo19} & Points & 88.70 & 95.30 \\
    3DCapsule\cite{YonghengCVPR2019} & Points & 88.90& -\\
    PointGrow\cite{YongbinCVPR2019} & Points & 85.80 & -\\
    PCGAN\cite{ChunLiangCVPR2019}& Points & 87.80 & -\\
    \hline
    Our & Points & \textbf{90.15} &\textbf{94.82}\\ % 90.153 % MN40 (90.68, 0.8791337) MN40 (90.7212, 87.9465)
    %Our(Two) & No & \textbf{88.33} &\textbf{83.53}\\ % 0.883306 class 0.835267
    %Our1(SN55) & No & 90.19&-\\
    %Our2(+SN55) & No & 91.25&-\\
    \hline
  \end{tabular}
\end{table}

\noindent \textbf{Shape segmentation.} We evaluate the local features learned by MAP-VAE for each point in shape segmentation. The ShapeNet part dataset~\cite{cvprpoint2017} is employed in this experiment, where point clouds in 16 shape classes are involved to train MAP-VAE with the same parameters in Table~\ref{table:comparison}.

\begin{figure}[tb]
  \centering
  % the following command controls the width of the embedded PS file
  % (relative to the width of the current column)
  %\includegraphics[width=.95\linewidth, bb=39 696 126 756]{figures/definition3.eps}
   \includegraphics[width=\linewidth]{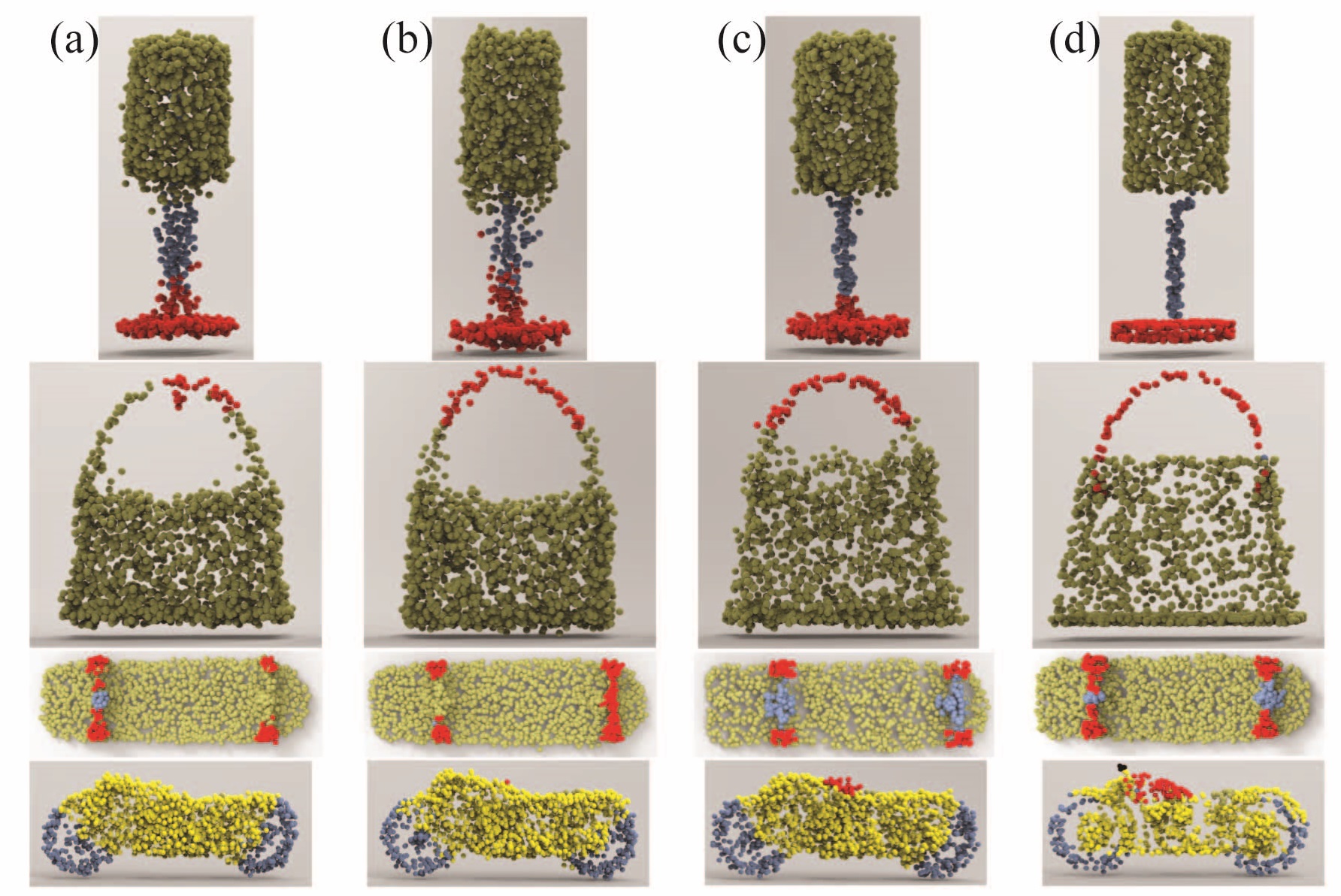}
  % replacing the above command with the one below will explicitly set
  % the bounding box of the PS figure to the rectangle (xl,yl),(xh,yh).
  % It will also prevent LaTeX from reading the PS file to determine
  % the bounding box (i.e., it will speed up the compilation process)
  % \includegraphics[width=.95\linewidth, bb=39 696 126 756]{sampleFig}
  %
  %
\caption{\label{fig:SegComp} Segmentation comparison between ``LGAN'' (a), ``LGAN1'' (b) in Table~\ref{table:segmentation} and MAP-VAE (c). The ground truth is shown in (d). A color in the same row represents a part class.}
\end{figure}

We first extracted the learned feature of each point from the second-last layer in the global decoder $\rm D$. Extracting the feature of each single point in the decoding procedure represents the ability of MAP-VAE to understand shapes locally at each point. Second, we map the ground truth label of each point to the reconstructed point cloud by voting 5 nearest labels for each reconstructed point. Third, we train a per-point softmax classifier under the training set, and test the classifier under the test set.

We use the same approach to obtain the results of the autoencoder in LGAN~\cite{PanosCVPR2018ICML}, and this autoencoder with a variational constraint. As the comparison shown in Table~\ref{table:segmentation}, our results significantly outperform ``LGAN'' and ``LGAN1'' in terms of both mIoU and classification accuracy. We further visualize the segmentation comparison on four cases in Fig.~\ref{fig:SegComp}. We find that the captured local geometry information helps MAP-VAE not only to reconstruct point clouds better, but also to learn more discriminative features for each point for better segmentation. Finally, we show 100 segmentation results in two challenging shape classes, respectively, i.e., airplane and chair, in Fig.~\ref{fig:Seg}. The consistent segmentation results also justify the good performance of MAP-VAE.

\begin{table*}
  \caption{The segmentation comparison among unsupervised 3D feature learning methods under ShapeNet part dataset. The metric is mIoU($\%$) and per-point classification accuracy($\%$) on points. $\alpha=0.01$,$\beta=1000$,$W=6$,$Z=128$.}%$\varepsilon=0.0004$,
  \label{table:segmentation}
  \centering
  \resizebox{\textwidth}{!}{
  \begin{tabular}{c|c|c|cccccccccccccccc}%llllllllll
    \hline
     &Methods& Mean &Aero& Bag& Cap& Car& Chair& Ear& Guitar & Knife & Lamp & Laptop & Motor & Mug & Pistol & Rocket & Skate & Table\\
    \hline		
     \multirow{3}{*}{\rotatebox{90}{mIoU}}&LGAN~\cite{PanosCVPR2018ICML}&57.04&54.13&48.67&62.55&43.24&68.37&58.34&74.27&68.38&53.35&82.62&18.60&75.08&54.70&37.17&46.71&66.39\\
     %PointNet++~\cite{nipspoint17}&58.8&61.8&59.2&65.0&52.1&81.0&58.4&75.3&56.6&63.0&92.6&22.0&48.7&52.3&22.6&62.8&67.4\\
     &LGAN1~\cite{PanosCVPR2018ICML}&56.28&52.16&57.85&62.66&42.01&67.66&52.25&75.37&68.63&49.07&81.52&19.20&75.43&54.34&35.09&41.48&65.73\\
     %&Ours($K=1$)&67.13&62.80&63.60&72.23&57.61&76.82&65.46&84.67&76.93&60.52&90.97&33.02&87.38&63.86&45.09&60.01&73.18\\
     %&Ours($K=3$)&67.49&63.47&63.56&72.48&57.84&77.27&69.66&84.67&76.92&60.20&90.85&34.12&87.25&63.77&45.22&58.69&73.93\\
     %&Ours($K=5$)&67.95&62.73&67.08&72.95&58.45&77.09&67.34&84.83&77.07&60.89&90.84&35.82&87.73&64.24&44.97&60.36&74.75\\
     &Ours&\textbf{67.95}&\textbf{62.73}&\textbf{67.08}&\textbf{72.95}&\textbf{58.45}&\textbf{77.09}&\textbf{67.34}&\textbf{84.83}&\textbf{77.07}&\textbf{60.89}&\textbf{90.84}&\textbf{35.82}&\textbf{87.73}&\textbf{64.24}&\textbf{44.97}&\textbf{60.36}&\textbf{74.75}\\
    \hline
    \multirow{3}{*}{\rotatebox{90}{ACC}}&LGAN~\cite{PanosCVPR2018ICML}&78.24&74.93&84.36&77.02&71.10&78.23&78.34&84.41&78.29&69.05&86.86&67.93&90.42&81.95&68.44&82.27&78.25\\
    &LGAN1~\cite{PanosCVPR2018ICML}&77.35&73.64&84.05&75.93&69.82&77.35&77.45&83.72&78.10&68.45&85.85&66.06&89.69&81.43&67.59&81.10&77.33\\
    %&Ours($K=1$)&87.27&83.30&93.30&85.68&82.85&86.87&88.23&93.13&86.57&79.05&94.98&76.34&98.82&90.40&77.97&92.85&86.05\\
    %&Ours($K=3$)&87.27&83.30&93.28&85.88&82.81&86.89&88.07&93.10&86.55&79.09&94.92&76.45&98.81&90.44&77.65&92.88&86.15\\
    %&Ours($K=5$)&87.45&83.50&93.79&86.12&83.28&87.03&88.08&93.15&86.66&79.31&94.89&77.37&98.86&90.51&77.14&93.21&86.25\\
    &Ours&\textbf{87.45}&\textbf{83.50}&\textbf{93.79}&\textbf{86.12}&\textbf{83.28}&\textbf{87.03}&\textbf{88.08}&\textbf{93.15}&\textbf{86.66}&\textbf{79.31}&\textbf{94.89}&\textbf{77.37}&\textbf{98.86}&\textbf{90.51}&\textbf{77.14}&\textbf{93.21}&\textbf{86.25}\\
    \hline
  \end{tabular}}
\end{table*}

\noindent \textbf{Shape generation.} Next we demonstrate how to generate novel shapes using the trained MAP-VAE. Here, we first sample a noise vector from the employed $Z$-dimensional unit Gaussian distribution, and then, convey the noise to the global decoder $\rm D$ in branch $\rm R$ in Fig.~\ref{fig:framework}.

\begin{figure}[tb]
  \centering
  % the following command controls the width of the embedded PS file
  % (relative to the width of the current column)
  %\includegraphics[width=.95\linewidth, bb=39 696 126 756]{figures/definition3.eps}
   \includegraphics[width=\linewidth]{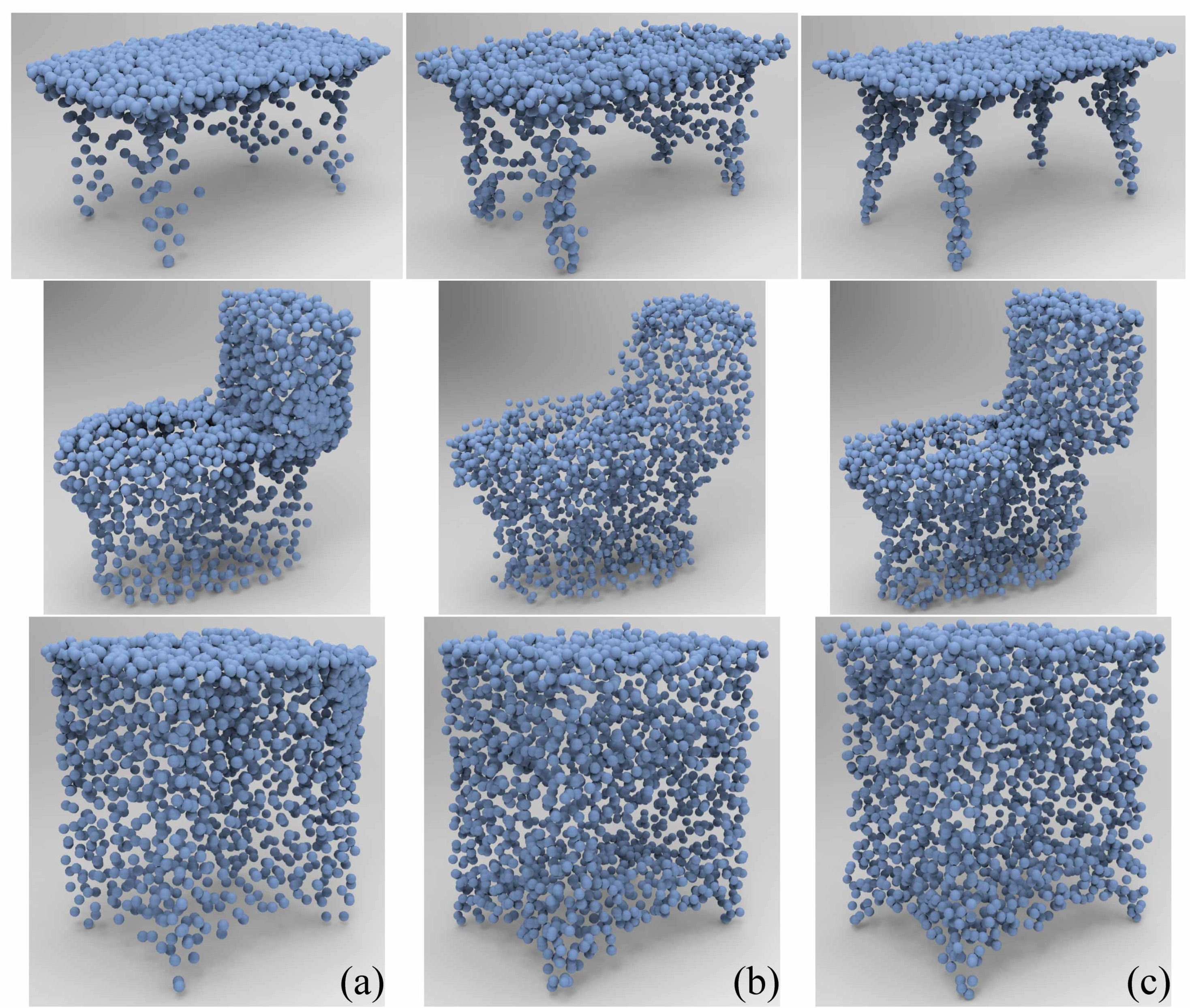}
  % replacing the above command with the one below will explicitly set
  % the bounding box of the PS figure to the rectangle (xl,yl),(xh,yh).
  % It will also prevent LaTeX from reading the PS file to determine
  % the bounding box (i.e., it will speed up the compilation process)
  % \includegraphics[width=.95\linewidth, bb=39 696 126 756]{sampleFig}
  %
  %
\caption{\label{fig:CenterComp} Compared to the three generated class centers under ModelNet10, MAP-VAE (in (c)) learns more local geometry details than ``LGAN'' (in (a)) and ``LGAN1'' (in (b)) in Table~\ref{table:segmentation}, due to the effective local self-supervision in half-to-half prediction.}
%Visual effect of half-to-half prediction under ModelNet10. Compared to the three class centers generated by ``LGAN'' (in (a)) and ``LGAN1'' (in (b)) in Table~\ref{table:segmentation}, MAP-VAE (in (c)) learns more local geometry details due to the effective local self-supervision in half-to-half prediction.
\end{figure}

\begin{figure}[tb]
  \centering
  % the following command controls the width of the embedded PS file
  % (relative to the width of the current column)
  %\includegraphics[width=.95\linewidth, bb=39 696 126 756]{figures/definition3.eps}
   \includegraphics[width=\linewidth]{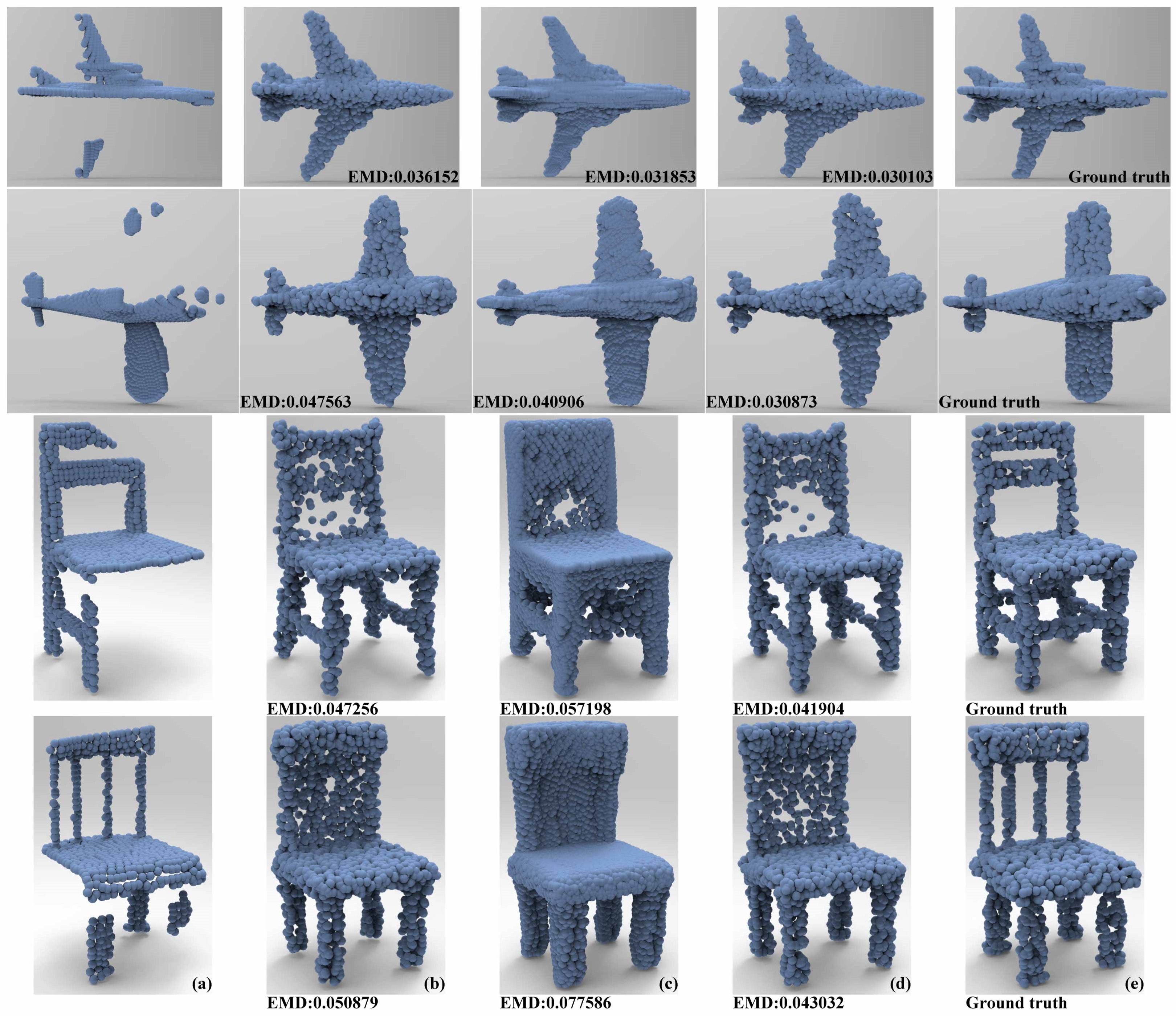}
  % replacing the above command with the one below will explicitly set
  % the bounding box of the PS figure to the rectangle (xl,yl),(xh,yh).
  % It will also prevent LaTeX from reading the PS file to determine
  % the bounding box (i.e., it will speed up the compilation process)
  % \includegraphics[width=.95\linewidth, bb=39 696 126 756]{sampleFig}
  %
  %
\caption{\label{fig:CompletionComp} Visual comparison with ``LGAN'' (in (b)) in Table~\ref{table:segmentation} and ``PCN-EMD'' (in (c)) in Table~\ref{table:completion} for the completion of partial point clouds (a). MAP-VAE (in (d)) completes more geometry details.
}
%Visually comparing MAP-VAE in (c) with ``LGAN'' (Table~\ref{table:segmentation}) in (b) and ``PCN-E'' (Table~\ref{table:completion}) in (c).

\end{figure}

\begin{figure*}[tb]
  \centering
  % the following command controls the width of the embedded PS file
  % (relative to the width of the current column)
  %\includegraphics[width=.95\linewidth, bb=39 696 126 756]{figures/definition3.eps}
   \includegraphics[width=\linewidth]{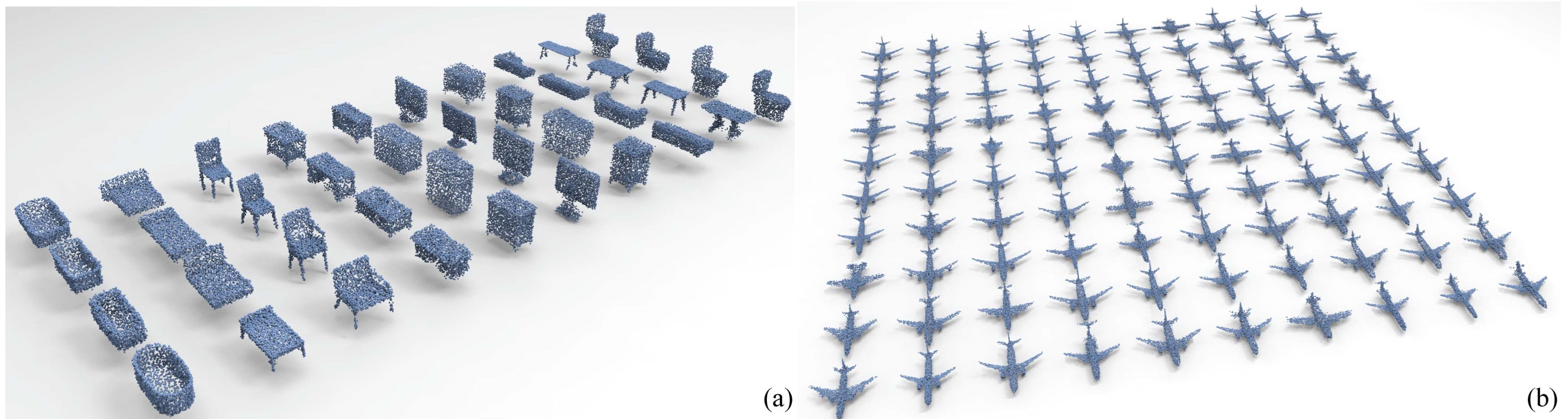}
  % replacing the above command with the one below will explicitly set
  % the bounding box of the PS figure to the rectangle (xl,yl),(xh,yh).
  % It will also prevent LaTeX from reading the PS file to determine
  % the bounding box (i.e., it will speed up the compilation process)
  % \includegraphics[width=.95\linewidth, bb=39 696 126 756]{sampleFig}
  %
  %
\caption{\label{fig:GenerateMN10} High fidelity novel shape generation by MAP-VAE trained under ModelNet10 in (a) and ShapeNet part dataset in (b).}
\end{figure*}

Using MAP-VAE trained under ModelNet10 in Table~\ref{table:comparison}, we generate some novel shapes in each of 10 shape classes in Fig.~\ref{fig:GenerateMN10} (a), where we sample 4 noise vectors around the feature center of each shape class to generate 4 shape class specific shapes. The generated point clouds are sharp and with high fidelity, where more local geometry details are learned. Moreover, we also observe high quality point clouds in the same shape class from MAP-VAE trained under ShapeNet part dataset in Table~\ref{table:segmentation}. We generate 100 airplanes using 100 noise vectors sampled around the feature center of the airplane class, as demonstrated in Fig.~\ref{fig:GenerateMN10} (b).

\begin{figure*}[tb]
  \centering
  % the following command controls the width of the embedded PS file
  % (relative to the width of the current column)
  %\includegraphics[width=.95\linewidth, bb=39 696 126 756]{figures/definition3.eps}
   \includegraphics[width=\linewidth]{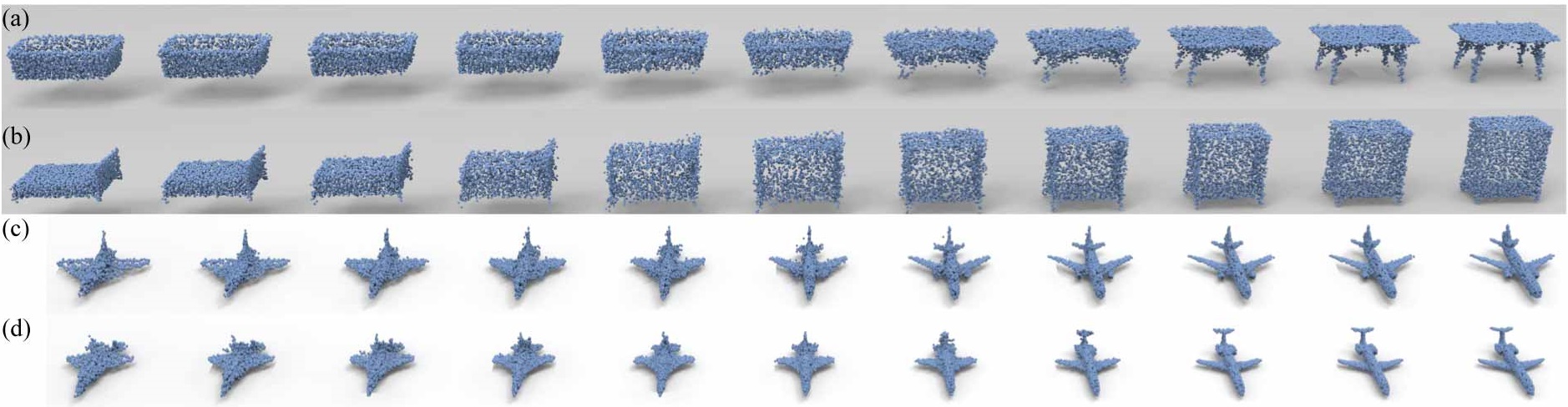}
  % replacing the above command with the one below will explicitly set
  % the bounding box of the PS figure to the rectangle (xl,yl),(xh,yh).
  % It will also prevent LaTeX from reading the PS file to determine
  % the bounding box (i.e., it will speed up the compilation process)
  % \includegraphics[width=.95\linewidth, bb=39 696 126 756]{sampleFig}
  %
  %
\caption{\label{fig:InterpolationMN10} We show shape interpolation results between two different shape classes under ModelNet10 in (a) and (b), and the shape interpolation results between two shapes in the same class under the ShapeNet part dataset in (c) and (d).}
\end{figure*}

\begin{table*}[tb]
  \caption{The completion comparison under airplane and chair classes in terms of EMD/point, $\alpha=0$,$\beta=1000$,$W=12$,$Z=0$.}%$\varepsilon=0.0004$,
  \label{table:completion}
  \centering
  \begin{tabular}{c|cccccc}%llllllllll
    \hline
     Class&EPN\cite{dai2017complete}&Folding\cite{YaoqingCVPR2018}&PCN-CD\cite{Yuan-2018-110219}&PCN-EMD\cite{Yuan-2018-110219}&LGAN\cite{PanosCVPR2018ICML}&Our\\
    \hline
     Airplane&0.061960&0.156438&0.046637&0.038752&0.033218&\textbf{0.032328}\\
     Chair&0.076802&0.297427&0.086787&0.068074&0.055908&\textbf{0.055696}\\% our:114.064771
    \hline		
\end{tabular}
\end{table*}

We further show the point clouds generated by the interpolated features between two feature centers of two shape classes, where the MAP-VAE trained under ModelNet10 in Table~\ref{table:comparison} is used. The interpolated point clouds are plausible and meaningful, and smoothly changed from one center to another center in Fig.~\ref{fig:InterpolationMN10} (a) and (b). Similar results can be observed by interpolations between two shapes in the same class in Fig.~\ref{fig:InterpolationMN10} (c) and (d), where the MAP-VAE trained under ShapeNet part dataset in Table~\ref{table:segmentation} is used.

Finally, we visually highlight the advantage of MAP-VAE by the point cloud generated at the feature center of a shape class. As demonstrated in Fig.~\ref{fig:CenterComp}, we compare MAP-VAE with the autoencoder in LGAN\cite{PanosCVPR2018ICML}, and this autoencoder with a variational constraint. Using the trained decoder of each method, we generate a point cloud from the feature center at each of the three shape classes. Compared to the three class centers in Fig.~\ref{fig:CenterComp} (a) and Fig.~\ref{fig:CenterComp} (b), MAP-VAE in Fig.~\ref{fig:CenterComp} (c) can generate point clouds with more local geometry details, such as sharp edges of parts.

\noindent\textbf{Point cloud completion. }MAP-VAE can also be used in point cloud completion, where we set $W=12$ and remove the KL loss for fair comparison. We evaluate our performance under the training and test sets of partial point clouds in two challenging shape classes in~\cite{dai2017complete}, i.e., airplane and chair, where we employ the complete point clouds in~\cite{cvprpoint2017} as ground truth. Since each partial point cloud has different number of points, we resample 2048 points to obtain the front and back halves. We compare with the state-of-the-art methods in Table~\ref{table:completion}. The lowest EMD distance shows that MAP-VAE outperforms all competitors. In addition, we also visually compare the completed point clouds in Fig.~\ref{fig:CompletionComp}, where MAP-VAE completes more local geometry details.

%Note that PCN employs 16384 points to represent a shape while MAP-VAE employs 2048 points, and under the same 150 shapes for testing, PCN only employs one partial point cloud scanned from the first view to obtain their evaluation result while MAP-VAE and LGAN employ all 8 partial point clouds for the evaluation.

\section{Conclusions}
We propose MAP-VAE for unsupervised 3D point cloud feature learning by jointly leveraging local and global self-supervision. MAP-VAE effectively learns local geometry and structure on point clouds from semantic local self-supervision provided by our novel multi-angle analysis. The outperforming results in various applications show that MAP-VAE successfully learns more discriminative global or local features for point clouds than state-of-the-art.%\footnote{This work was supported by National Key R\&D Program of China (2018YFB0505400) and NSF under award number 1813583.}

%\section*{Acknowledgments}
%This work was supported by National Key R\&D Program of China (2018YFB0505400) and NSF under award number 1813583.
%~\\
%\noindent\textbf{Acknowledgments} This work was supported by National Key R\&D Program of China (2018YFB0505400) and NSF under award number 1813583.
%\footnote{This work was supported by National Key R\&D Program of China (2018YFB0505400) and NSF under award number 1813583.}

{\small
\bibliographystyle{ieee}
\bibliography{paper}
}

\end{document}